# An Integrated Framework for Learning and Reasoning

**Christophe G. Giraud-Carrier** CGC@COMPSCI.BRISTOL.AC.UK
*Department of Computer Science, University of Bristol*
*Bristol, BS8 1TR U.K.*

**Tony R. Martinez** MARTINEZ@CS.BYU.EDU
*Department of Computer Science, Brigham Young University*
*Provo, UT 84602 U.S.A.*

## Abstract

Learning and reasoning are both aspects of what is considered to be intelligence. Their studies within AI have been separated historically, learning being the topic of machine learning and neural networks, and reasoning falling under classical (or symbolic) AI. However, learning and reasoning are in many ways interdependent. This paper discusses the nature of some of these interdependencies and proposes a general framework called FLARE, that combines inductive learning using prior knowledge together with reasoning in a propositional setting. Several examples that test the framework are presented, including classical induction, many important reasoning protocols and two simple expert systems.

## 1. Introduction

Induction and deduction are both underlying processes in intelligent agents. Induction "involves intellectual leaps from the particular to the general" (D'Ignazio & Wold, 1984). It plays an important part in knowledge acquisition or learning. D'Ignazio and Wold (1984) claim that indeed, "All the laws of nature were discovered by inductive reasoning." Deduction is a form of reasoning with and about acquired knowledge. It typically does not result in the generation of new facts, rather it establishes cause-effect relationships between existing facts. Deduction may be applied *forward* by seeking the consequences of certain existing hypotheses or *backward* to discover the necessary conditions for the achievement of certain goals. Despite their differences, induction and deduction are strongly interrelated. The ability to reason about a domain of knowledge is often based on rules about that domain, that must be acquired somehow; and the ability to reason can often guide the acquisition of new knowledge or learning.

Inductive learning has been the subject of much research leading to the design of a variety of algorithms (e.g., Clark & Niblett, 1989; Michalski, 1983; Quinlan, 1986; Salzberg, 1991). In general, inductive learning systems generate classification rules from examples. Typically, the system is first presented a set of examples (objects, situations, etc.), also known as a *training set*. Examples are usually expressed in the attribute-value language and represent recorded instances of attribute-value pairs together with their corresponding classification. The system's goal is then to discover sets of sufficient critical features or rules that properly classify the examples of the training set (convergence) and adequately extend to previously unseen examples (generalization).

Though machines are still a far cry from matching human qualitative inductive leaps, inductive learning systems have proven useful over a wide range of applications in medicine (breast cancer, hepatitis detection), banking (credit screening), defense (mine-rock discrimination), botany (iris variety identification, venomous mushroom detection) and others (Murphy & Aha, 1992).

The study of deductive reasoning goes at least as far back as the early Greek philosophers, such as Socrates and Aristotle. Its formalization has given rise to a variety of logics, from propositional to first-order predicate logic to default logic to several non-monotonic extensions. Many of these logics have been successfully implemented in artificial systems (e.g., PROLOG, expert systems). They typically consist of a pre-encoded knowledge or rule base, a given set of facts (identified as either causes or consequences) and some *inference engine*. The inference engine carries out the deductive process using the rules in the rule base and the facts it is provided. Several of these systems have been successfully used in various domains, such as medical diagnosis (Clancey & Shortliffe, 1984) and geology (Duda & Reboh, 1984).

One of the greatest challenges of current deductive systems is *knowledge acquisition*, that is, the construction of the rule base. Typically, the rule base is generated as domain knowledge is extracted from human experts and carefully engineered into rules. Knowledge acquisition is a tedious task that presents many difficulties both practically and theoretically. If a sufficiently rich training set can be obtained, then inductive learning may be used effectively to complement the traditional approach to knowledge acquisition. Indeed, a system's knowledge base can be constructed from both rules encoded a priori and rules generated inductively from examples. In other words, rules and examples need not be mutually exclusive. The strong knowledge principle (Waterman, 1986) and early work on bias (Mitchell, 1980) suggest the need for prior knowledge. Rules supplied a priori are one simple form of prior knowledge that has been used successfully in several inductive systems (e.g., Giraud-Carrier & Martinez, 1993; Ourston & Mooney, 1990). Similarly, proposals have been made to enhance deductive systems with learning capabilities (e.g., Haas & Hendrix, 1983; Rychener, 1983).

It is these authors' contention that the study of the interdependencies between learning and reasoning and the subsequent integration of induction and deduction into unified frameworks may lead to the development of more powerful models. This paper describes a system, called FLARE (Framework for Learning And REasoning), that attempts to combine inductive learning using prior knowledge together with reasoning. Induction and deduction in FLARE are carried out within the confines of non-recursive, propositional logic. Learning is effected incrementally as the system continually adapts to new information. Prior knowledge is given by a teacher in the form of rules. Within the context of a particular inductive task, these rules may serve to produce useful learning biases. Simple defaults combined with learning capabilities enable FLARE to exhibit reasoning that is normally considered non-monotonic.

The paper is organized as follows. Section 2 presents FLARE and argues the validity of the unified framework. FLARE's representation language is described and the algorithms employed in learning and reasoning are detailed. Section 3 reports experimental results on classical datasets, a number of "well-designed" reasoning protocols and several other applications, including two simple expert systems. Some of the limitations of the system are also described. Section 4 discusses related work in induction and deduction. Finally, section 5 concludes the paper by summarizing the results and discussing further research.

## 2. FLARE - A Framework for Learning and Reasoning

In this section, FLARE's learning and reasoning mechanisms are detailed. A description and discussion of FLARE's representation language are given first in Section 2.1, along with some useful definitions and a simple, practical example that will serve as a running example throughout the paper. Sections 2.2 to 2.5 then follow a top-down approach to the description of FLARE.

### 2.1 FLARE's Representation Language

FLARE's representation language is an instance of the attribute-value language (AVL). In FLARE, attributes may range over nominal domains and bounded linear domains, including closed intervals of continuous numeric values. The basic elements of knowledge in AVL are vectors defined over the cross-product of the domains of the attributes. The components of a vector specify a value for each attribute. The following simple extension is made to AVL. If $A$ is an attribute and $D$ is the domain of $A$, then $A$ takes on values from $D \cup \{\star, ?\}$. The special symbols $\star$ and ? stand for *don't-care* and *don't-know*, respectively.

The semantics associated with $\star$ and ? are different. An attribute whose value is $\star$ is one that is known (or assumed) to be irrelevant in the current context, while an attribute whose value is ? may be relevant but its actual value is currently unknown. The $\star$ symbol allows the encoding of rules, while the ? symbol accounts for missing attribute values in real-world observations.

#### 2.1.1 First-Order to Attribute-Value Translation

Since learning and reasoning tasks are often expressed in English with simple, direct counterparts in the classical first-order logic language (FOL), it is necessary for FLARE to translate FOL clauses into their AVL equivalent. AVL is clearly not as expressive as FOL, so that FLARE has some inherent limitations. For the purposes of this discussion, let predicates of the form $p(x)$ and $p(x, C)$ where $C$ is a constant be called *avl-predicates*. Then, the FOL clauses that can be translated into AVL are of two kinds:

1. **ground facts:** $p(C)$ or $\neg p(C)$ where $C$ is a constant (e.g., block($A$)).
2. **simple implications:** $(\forall x)P(x) \Rightarrow q(x)$ where $P(x)$ is a conjunction of avl-predicates and $q(x)$ is, without loss of generality, a single, possibly negated avl-predicate (e.g., block($x$) $\wedge$ weight($x, heavy$) $\Rightarrow$ $\neg$on_table($x$)).

All clauses involve at most one universally quantified variable and are thus essentially non-recursive, propositional clauses. Despite its restricted language, FLARE effectively handles a significant range of applications. Moreover, AVL accounts for simple, efficient matching mechanisms and lends itself naturally to many inductive learning problems as witnessed by its use in many successful learning systems (Clark & Niblett, 1989; Michalski, 1983; Quinlan, 1986).

FOL statements of the aforementioned forms are translated in a straightforward way into an equivalent symbolic-valued AVL representation, as shown in Figure 1. A similar transformation has been proposed in the context of ILP (Džeroski, Muggleton, & Russell, 1993). Like FLARE, some ILP systems, such as LINUS (Lavrač, Džeroski, & Grobelnik,

1. **Attribute definition:** For each avl-predicate, create a matching Boolean (for $p(x)$) or multi-valued (for $p(x, C)$) attribute. If there are ground facts, create a multi-valued attribute, called *label*, whose values are those of the constants.
2. **Vector definition:** For each implication, create a matching vector where attributes corresponding to premise and conclusion have their appropriate value and all other attributes are set to $\star$. For each ground fact, create a matching vector where the value of *label* is that of the constant and the attribute corresponding to the predicate has its appropriate value. Tag the attribute corresponding to the conclusion.

Figure 1: FOL to AVL Transformation

| FOL | AVL | | |
|---|---|---|---|
| | Rep (b) | Qua (b) | Pac (b) |
| Republican($x$)$\Rightarrow$ $\neg$Pacifist($x$) | 1 | $\star$ | $0_T$ |
| Quaker($x$)$\Rightarrow$Pacifist($x$) | $\star$ | 1 | $1_T$ |

Table 1: Nixon Diamond (Reiter & Griscuolo, 1981)

1991), first map ILP problems to propositional learning problems and then rely on attribute-based learning.

The creation of attribute *label* in step 2 stems from the fact that ground facts of the form $p(C)$ can be rewritten as simple implications of the form $label(x, C) \Rightarrow p(x)$. Notice how the attributes whose values are $\star$ in a vector correspond exactly to those predicates that do not appear in the premise of the corresponding FOL clause. The attribute corresponding to $q(x)$ has different usages. It functions as a conclusion during forward chaining and as a target classification during inductive learning. In some cases, it can also be used as a goal. To avoid unnecessary confusion, the attribute corresponding to $q(x)$ is simply referred to as the *target-attribute*. The values of the target-attribute are subsequently tagged with the subscript $T$. The translation from FOL to AVL is currently performed manually.

It is clear that as the number of predicates increases, so does the size of the vectors. Since all vectors are of the same size and many of them may only have values set for a relatively small number of their attributes, this may result in large memory requirements, as well as in an increase of execution time of operations on vectors. When there are predicates that qualify different values of the same concept (e.g., red($x$), yellow($x$), for color), it is possible to limit the size of the vectors by translating such predicates into a single multi-valued attribute (e.g., color($x, V$), where $V$ is a constant: red, yellow, etc.). This is particularly useful for the conclusion part $q(x)$ when it corresponds to a classification for $x$.

Tables 1 through 4 contain four simple examples that demonstrate the transformation. Each derived attribute in the AVL column is followed by its type (b for Boolean, m for multi-valued). Table 1 shows the Nixon Diamond, a classical example of conflicting defaults. Informally, the Nixon Diamond states that Republicans are typically not pacifist but Quakers are typically pacifist. The conflict then arises as one asserts that Nixon is both a Republican and a Quaker. Table 2 contains assertions about animals and their ability to fly. It states that animals normally do not fly, birds are typically flying animals and penguins are birds that do not fly. Table 3 shows statements regarding eyes and their fitness for lenses. Finally, Table 4 contains some facts about a simple blocks world.

| FOL | AVL | | | |
|---|---|---|---|---|
| | Ani (b) | Bir (b) | Pen (b) | Fly (b) |
| Animal($x$)⇒ ¬Fly($x$) | 1 | $\star$ | $\star$ | $0_T$ |
| Bird($x$)⇒Animal($x$) | $1_T$ | 1 | $\star$ | $\star$ |
| Bird($x$)⇒Fly($x$) | $\star$ | 1 | $\star$ | $1_T$ |
| Penguin($x$)⇒Bird($x$) | $\star$ | $1_T$ | 1 | $\star$ |
| Penguin($x$)⇒ ¬Fly($x$) | $\star$ | $\star$ | 1 | $0_T$ |

Table 2: Flying or Not Flying (Lifschitz, 1988)

| FOL | AVL | | |
|---|---|---|---|
| | Tpr (m) | Eye (m) | Fit (b) |
| Tear-prod-rate($x$,low)⇒Eyes($x$,dry) | low | $\text{dry}_T$ | $\star$ |
| Eyes($x$,dry)⇒ ¬Fit($x$) | $\star$ | dry | $0_T$ |

Table 3: Fitting Lenses

#### 2.1.2 Examples vs. Precepts vs. Rules

Informally, the problem of supervised learning may be described as follows. Given (1) a set of categories, (2) for each category, a set of instances of "objects" in that category and (3) optional prior knowledge, produce a set of rules sufficient to place objects in their correct category. In AVL, instances consist of sets of attribute-value pairs or vectors, describing characteristics of the objects they represent, together with the object's category. In this context, the category is a target-attribute.

An *example* is a vector in which all attributes are set to either ? or one of their possible values. A *rule* is a vector in which some of the attributes have become $\star$ as a result of generalization during inductive learning. A *precept* is similar to a rule but, unlike a rule, it is not induced from examples. Precepts are either given by a teacher or deduced from general knowledge relevant to the domain under study. In the context of a given rule or precept, the $\star$ attributes have no effect on the value of the category. Precepts and rules thus *represent* several examples. For instance, let $p = (\star, 1, 0, 0_T)$ be a precept, where all attributes range over the set {0,1,2}. Then $p$ represents the three examples: $(0, 1, 0, 0_T)$, $(1, 1, 0, 0_T)$ and $(2, 1, 0, 0_T)$.

The distinction between rules and precepts is limited to learning. In reasoning, all vectors (including examples that do not generalize) are rules. In FLARE, rules are formed by dropping conditions (Michalski, 1983), that is, under certain circumstances (see Section 2.4.2), one attribute is set to $\star$. Precepts, on the other hand, are rules encoded a priori. They reflect some high-level knowledge (or common sense) about the real-world. A precept "suggests something advisory and not obligatory communicated typically through teaching" (Webster's Dictionary).

#### 2.1.3 Running Example

To illustrate the above definitions and algorithms of the following sections, a final example of the transformation is constructed, based on the *mediadv* knowledge base (Harmon & King, 1985). This purposely simple example will serve as a running example throughout the paper. A discussion of the complete *mediadv* knowledge base is in Section 3.5. Here, two conditions (i.e., instructional feedback and presentation modification) are left out and only a few of the original rules are used. Table 5 contains the informal English version of the knowledge used (with reference to the rules of *mediadv* it was generated from when applicable) and its corresponding translation into AVL vectors.

| FOL | AVL | | | |
|---|---|---|---|---|
| | Lab (m) | Blk (b) | Hvy (b) | OnT (b) |
| block($A$) | $A$ | $1_T$ | $\star$ | $\star$ |
| block($B$) | $B$ | $1_T$ | $\star$ | $\star$ |
| heavy($A$) | $A$ | $\star$ | $1_T$ | $\star$ |
| heavy($B$) | $B$ | $\star$ | $1_T$ | $\star$ |
| block($x$) $\wedge$ heavy($x$) $\Rightarrow$ on_table($x$) | $\star$ | 1 | 1 | $1_T$ |
| $\neg$on_table($A$) | $A$ | $\star$ | $\star$ | $0_T$ |

Table 4: Simple Blocks World, adapted from (Lifschitz, 1988)

Let $KB$ be the resulting set of vectors. The attributes are given in the order: situation, stimulus-situation, response, appropriate-response, stimulus-duration, training-budget and media. Note that all the attributes are nominal. The symbolic values used in the English statements are transformed into equivalent nominal values in the vectors. Hence, for example, the first statement gives rise to a vector in which the attribute situation is set to 0 (the corresponding nominal value of schematics for this attribute), and the target-attribute stimulus-situation is set to 0 (the corresponding nominal value of symbolic for this attribute).

The top goal is for the system to suggest the most effective media for training, based on four conditions: situation, response, stimulus-duration, and training-budget. Note that the attributes stimulus-situation and appropriate-response can be used as subgoals in reaching the final conclusion. Vectors $v_{13}$ to $v_{17}$ are examples since all of the conditions have set values. They are not part of the original *mediadv* knowledge base but are added to exercise important features of the algorithms. As $KB$ is given, all vectors of $KB$ with condition attributes set to $\star$ are precepts rather than rules. For instance, $v_7$ and $v_{12}$ are precepts. Then, the term rule applies to new generalizations, induced by FLARE from $KB$. For instance, $v_8'$ (see Section 2.4.3) is a rule.

## 2.2 Algorithmic Overview

FLARE is a self-adaptive, incremental system. It uses domain knowledge and empirical evidence to construct and maintain its knowledge base. FLARE's knowledge base is interpreted as a "best so far" set of rules for coping with the current application. In that sense, FLARE follows the scientific approach to theory formation/revision: available prior knowledge and experience produce a "theory" that is updated or refined continually by new evidence.

FLARE involves three main functions whose definitions and high-level algorithmic interactions are given in Figure 2. The details of each function's implementation are given in the following sections. An intuitive overview is presented here.

| | English Statements | | Equivalent AVL Vectors | | | | | | | | |
|---|---|---|---|---|---|---|---|---|---|---|---|
| If | situation = schematics | (Rule3) | $v_1$ | = | 0 | $0_T$ | $\star$ | $\star$ | $\star$ | $\star$ | $\star$ |
| Then | stimulus-situation = symbolic | | | | | | | | | | |
| If | situation = conversation | (Rule4) | $v_2$ | = | 1 | $1_T$ | $\star$ | $\star$ | $\star$ | $\star$ | $\star$ |
| Then | stimulus-situation = verbal | | | | | | | | | | |
| If | situation = photograph | (Rule2) | $v_3$ | = | 2 | $2_T$ | $\star$ | $\star$ | $\star$ | $\star$ | $\star$ |
| Then | stimulus-situation = pictorial | | | | | | | | | | |
| If | response = observing or | (Rule5) | $v_4$ | = | $\star$ | $\star$ | 0 | $0_T$ | $\star$ | $\star$ | $\star$ |
| | response = thinking | | $v_5$ | = | $\star$ | $\star$ | 1 | $0_T$ | $\star$ | $\star$ | $\star$ |
| Then | appropriate-response = covert | | | | | | | | | | |
| If | response = emoting | (Rule10) | $v_6$ | = | $\star$ | $\star$ | 2 | $1_T$ | $\star$ | $\star$ | $\star$ |
| Then | appropriate-response = affective | | | | | | | | | | |
| If | stimulus-situation = verbal and | (Rule13) | $v_7$ | = | $\star$ | 1 | $\star$ | 0 | 0 | $\star$ | $2_T$ |
| | appropriate-response = covert and | | | | | | | | | | |
| | stimulus-duration = brief | | | | | | | | | | |
| Then | media = lecture | | | | | | | | | | |
| If | stimulus-situation = verbal or | (Rule14) | $v_8$ | = | $\star$ | 1 | $\star$ | 0 | 0 | 1 | $3_T$ |
| | stimulus-situation = symbolic or | | $v_9$ | = | $\star$ | 0 | $\star$ | 0 | 0 | 1 | $3_T$ |
| | stimulus-situation = pictorial and | | $v_{10}$ | = | $\star$ | 2 | $\star$ | 0 | 0 | 1 | $3_T$ |
| | appropriate-response = covert and | | | | | | | | | | |
| | stimulus duration = brief and | | | | | | | | | | |
| | training-budget = medium | | | | | | | | | | |
| Then | media = lecture-with-slides | | | | | | | | | | |
| If | stimulus-situation = verbal and | (Rule16) | $v_{11}$ | = | $\star$ | 1 | $\star$ | $\star$ | 0 | $\star$ | $0_T$ |
| | stimulus-duration = brief | | | pty = 1 | | | | | | | |
| Then | media = role-play-w/verbal-feedback | | | | | | | | | | |
| If | stimulus-situation = verbal and | (Rule17) | $v_{12}$ | = | $\star$ | 1 | $\star$ | 1 | $\star$ | $\star$ | $1_T$ |
| | appropriate-response = affective | | | pty = 3 | | | | | | | |
| Then | media = role-play-w/video-feedback | | | | | | | | | | |
| If | situation = conversation and | | $v_{13}$ | = | 1 | $\star$ | 0 | $\star$ | 0 | 1 | $2_T$ |
| | response = observing or | | $v_{14}$ | = | 1 | $\star$ | 1 | $\star$ | 0 | 1 | $2_T$ |
| | response = thinking and | | | | | | | | | | |
| | stimulus-duration = brief and | | | | | | | | | | |
| | training budget = medium | | | | | | | | | | |
| Then | media = lecture | | | | | | | | | | |
| If | situation = photograph and | | $v_{15}$ | = | 2 | $\star$ | 2 | $\star$ | 1 | 0 | $0_T$ |
| | response = emoting and | | | | | | | | | | |
| | stimulus-duration = persistent and | | | | | | | | | | |
| | training-budget = small | | | | | | | | | | |
| Then | media = role-play-w/verbal-feedback | | | | | | | | | | |
| If | situation = photograph and | | $v_{16}$ | = | 2 | $\star$ | 2 | $\star$ | 1 | 0 | $2_T$ |
| | response = emoting and | | | | | | | | | | |
| | stimulus-duration = persistent and | | | | | | | | | | |
| | training-budget = small | | | | | | | | | | |
| Then | media = lecture | | | | | | | | | | |
| If | situation = photograph and | | $v_{17}$ | = | 2 | $\star$ | 2 | $\star$ | 1 | 0 | $0_T$ |
| | response = emoting and | | | | | | | | | | |
| | stimulus-duration = persistent and | | | | | | | | | | |
| | training-budget = small | | | | | | | | | | |
| Then | media = role-play-w/verbal-feedback | | | | | | | | | | |

Table 5: Simple *KB* Running Example

**DEFINITION**

- **Function:** Generate-Precepts
  - Input: a set of general rules, a set of facts and one designated target-attribute.
  - Output: one or more precepts.
- **Function:** Reasoning
  - Input: the current knowledge base, a set of facts encoded in a vector $v$, one designated target-attribute and optionally, the target value of the target-attribute.
  - Output: a vector $v^+$ equal to $v$ together with further facts deduced from $v$, including a derived value for the target-attribute.
- **Function:** Adapting
  - Input: the current knowledge base, the vector $v^+$ output by function Reasoning and the target value of the target-attribute.
  - Output: updated knowledge base.

**IMPLEMENTATION**

1. **Preprocessing:** Perform Generate-Precepts
2. **Main loop:** For each vector presented to the system
   (a) Perform Reasoning
   (b) If there is a target value for the target-attribute, perform Adapting

Figure 2: FLARE - Algorithmic Overview

Conceptually, FLARE's execution consists of two phases. In the preprocessing phase, FLARE uses prior knowledge in the form of general rules that may be viewed as encoding "commonsense" knowledge. Using deduction from given facts, domain-specific precepts are generated as an instantiation of the general knowledge to the domain at hand. Section 2.5 details the Generate-Precepts function. The need for generating and explicitly encoding precepts as individual vectors in such a preprocessing phase arises because FLARE's inductive mechanisms take place at the vector level. Thus, even though it is always possible to deduce them from the general knowledge, precepts are most useful in induction when they are made explicit.

In normal processing, FLARE executes an, at least conceptually, infinite loop. Steps (a) and (b) are executed every time new information (in the form of AVL vectors) is presented to the system. In step (a), FLARE reasons from the "facts" provided by the input vector and the rules found in the current knowledge base. Rule-based reasoning and similarity-based reasoning are combined as discussed in Section 2.3 to derive a value for the target-attribute, as well as other attributes along the forward chain to the conclusion. In step (b), FLARE adapts its current knowledge base. Because FLARE is a supervised learner, it can only adapt when a target value for the target-attribute is explicitly given as part of the information presented. The combination of steps (a) and (b) is referred to as *learning*. Section 2.4 details the Adapting function.

Note that reasoning based upon available knowledge prior to adapting is plausible. Even when available information is insufficient and/or incomplete, humans often attempt to make a tentative decision and get corrected if necessary. At any one time, the decision made represents a kind of "best guess" given currently available information. The more (correct) information becomes available, the more accurate decisions become.

## 2.3 FLARE's Reasoning

FLARE implements a simple form of rule-based reasoning combined with similarity-based reasoning, similar to CONSYDERR (Sun, 1992). Sun has argued that such a combination effectively decreases the system's susceptibility to brittleness (Sun, 1992). In particular, in the absence of applicable rules or when information is incomplete, FLARE relies on similarity with previously encountered situations to make useful predictions. Others have also argued that analogy is a necessary condition for commonsense reasoning and the subsequent overcoming of brittleness (Minsky & Riecken, 1994; Wollowski, 1994). Section 2.3.1 shows how the notion of Clark's completion (1978) can be applied to inductively learned rules and exploited by similarity-based reasoning to generate new rules. Sections 2.3.2 to 2.3.7 describe and illustrate FLARE's reasoning mechanisms.

### 2.3.1 Completion

Inductively learned rules of the form $(\forall x)P(x) \Rightarrow q(x)$, where $P$ is a conjunction of avl-predicates, are essentially classification rules or definitions that establish relationships between features, captured by $P(x)$, and concepts, expressed by $q(x)$. In keeping with the classical assumption that what is not known by a learning system is false by default, inductively generated rules lend themselves naturally to the completion principle proposed by Clark (1978). That is, classification rules become "if and only if" statements, i.e., $P(x) \Leftrightarrow q(x)$. Hence, under completion, if $q(x)$ is known to be true, then it is possible to conclude that $P(x)$ is true.

Clearly, completion does not apply to all rules. Inductively learned rules are inherently definitional as they essentially encode a concept's description in terms of a set of features. Other rules, such as those relating concepts at the same relative cognitive level, are not definitional. For example, given that birds are animals and that some $x$ is an animal, it does not follow that $x$ is a bird. Note that, in addition to inductively learned rules, definitions may be given to FLARE as prior knowledge.

The completion principle is particularly useful when it interacts with similarity-based reasoning to generate new rules, as shown in the following derivation.

- **Hypotheses:**
    1. $(\forall x)P(x) \Rightarrow q(x)$, which may be completed.
    2. $(\forall x)P'(x) \Rightarrow q'(x)$.
    3. $P \cap P' \neq \emptyset$ (i.e., $P$ and $P'$ have some attributes in common).
    4. $q(x)$ is true.
- **Derivation:**
    1. $q(x)$ from hypothesis 4.

2. $P(x)$ from completion applied to hypothesis 1.
3. $q'(x)$ from similarity-based reasoning using hypotheses 2 and 3.

A new implication between concepts, namely $q(x) \Rightarrow q'(x)$, is thus generated. Though FLARE is capable of deriving $q'(x)$ from $q(x)$, it does not actually store the new implication in its knowledge base.

The following example adapted from (Collins & Michalski, 1989) illustrates the use of the above derivation. Assume that the system has learned a description of the Chaco area in terms of a set $G$ of geographical conditions (i.e., $G(x) \Rightarrow$area$(x, theChaco)$). Furthermore, assume that the system knows a rule that encodes a set of conditions $C$ sufficient for the raising of cattle (i.e., $C(x) \Rightarrow$raise$(x, cattle)$) and $C$ is such that $C$ and $G$ share a number of conditions. If the system is now told that the area of interest is the Chaco, it first deduces by completion that the conditions in $G$ are met and then, by taking advantage of the similarity between $G$ and $C$, the system concludes that cattle may be raised in the Chaco. Note that the level of confidence in the conclusion depends upon the amount of similarity.

In FLARE, the representation is extended and a *definition indicator* is tagged to those statements that may be completed (i.e., prior definitions, inductively learned classifications). Note that, though somewhat cumbersome, this extension is needed since FLARE does not physically separate concepts and the features used to describe them. CONSYDERR on the other hand provides natural support for the dichotomy. FLARE's representation makes learning more readily applicable and preserves consistency with previously developed models. At this point, the issue of achieving both the dichotomy and easy learning remains open.

#### 2.3.2 FLARE's Reasoning Function

Deduction in FLARE is applied forward. Hence, facts must be provided so as to initiate reasoning. These facts are coded into a vector in which attributes whose values are known are accordingly set, while all other attributes are ? (i.e., don't-know). One attribute is designated as the target-attribute and, if known, its value is also provided. FLARE then uses the rules of its knowledge base and the facts to derive a value for the target-attribute. The Reasoning function is shown in Figure 3. Note that in this discussion, the current knowledge base is assumed to be non-empty. If the knowledge base is empty, the system cannot deduce anything other than ?.

Step (1) applies completion first. FLARE finds all asserted (i.e., neither $\star$ nor ?) attributes of $v$ that are target-attributes of definitions in the current knowledge base. If any such attribute is found, and for all of them, completion is applied by "copying" into $v$ all asserted attributes of the corresponding definitions that are ? in $v$. The following two issues must be addressed by FLARE in implementing completion.

1. Since some attributes may be involved in the definitions of more than one target-attribute or concept, it follows that there may be more than one values to be copied into a given attribute when completing these definitions.

2. Since FLARE's concepts and rules consist of sets of vectors, where each vector is a conjunction and all the vectors sharing the same target-attribute form a disjunction, it follows that some definitions may be disjunctive as well.

**DEFINITION**

- Input: the current knowledge base, a set of facts encoded by a vector $v$, one designated target-attribute and optionally, the target value of the target-attribute.
- Output: a vector $v^+$ equal to $v$ together with further facts deduced from $v$, including a value for the target-attribute.

**IMPLEMENTATION**

1. **Completion:** For each asserted attribute $a$ of $v$ other than the target-attribute, if $a$ is the target-attribute of a definition $d$ and their values are equal, then copy all asserted attributes of $d$ that are ? in $v$, into $v$.
2. **Forward chaining:** If $v$'s target-attribute has not been asserted
   (a) Repeat until no new attribute of $v$ has been asserted
      i. Let $w = v$.
      *(* create a temporary copy of v *)*
      ii. For each non-asserted attribute $a$ of $v$ other than the target-attribute, if a rule can be applied to $v$ to assert $a$, then apply it by asserting $a$ in $w$.
      *(* based on v, assert all possible attributes (other than the target-attribute) in w *)*
      iii. Let $v = w$.
      *(* copy result back into v for next level of inference *)*
   (b) If a rule can be applied to assert $v$'s target-attribute, then apply it. Otherwise, perform similarity-based assertion.

Figure 3: Function Reasoning

The current implementation resolves these two issues as follows. In the first case, potential conflicts are resolved simply by giving precedence to the first copy made (which depends upon the order in which asserted attributes are processed). In the second case, FLARE simply chooses one of the defining conjunctions at random and applies completion to it. Other mechanisms (e.g., apply to all, select a winner based on some criteria, etc.) are the topic of further research.

Completion causes further information (in the form of asserted attributes) to be gained, thus improving the chance of reaching a goal. Indeed, the purpose of step (1) is two-fold. First, completion allows the system to reach goals that are not otherwise achievable by existing rules. Second, even if the top goal is not achieved directly by completion, further reasoning to achieve it is enhanced as described in Section 2.3.1.

When the target-attribute has not been asserted by completion, step (2) pursues the reasoning process using forward chaining. As mentioned above, $v$ has a single target-attribute, corresponding to the final goal to achieve. However, at any given time, any one of the (yet) non-asserted[1] attributes of $v$ may be designated as a subgoal that may

1. These are either $\star$ or ?. They are $\star$ when precepts and rules with differing premises and conclusions are used. In such cases, it is not clear until reasoning whether they are true don't-cares or only don't-knows.

be useful (or necessary) in reaching the final conclusion. Step (2)(a) is the heart of the reasoning process. Each execution of step (2)(a)(ii) corresponds to the achievement of all possible subgoals at a given depth in the inference process. Each iteration uses knowledge acquired in the previous iteration to attempt to derive more new conclusions using existing rules. Step (2)(b) concludes the reasoning phase by asserting the target-attribute.

Notice that the target-attribute is always asserted, either by rule application or similarity-based assertion. Hence, FLARE always reaches a conclusion. In the worst case, when there is no information about the target-attribute in the current knowledge base, the value derived for the conclusion must clearly be ?. In all other cases, the validity and accuracy of the derived conclusion depend upon available information. The accuracy or confidence level may be computed in a variety of ways from information about static priorities, dynamic priorities, covers and counters (see Section 2.4).

The two complementary mechanisms used in asserting the target-attribute (i.e., rule application and similarity-based assertion) are described in the next two sections. They apply sequentially. If a rule exists that can be applied, then it is applied. Otherwise, similarity-based reasoning takes effect.

Finally, note that information regarding the way the goal is achieved could be displayed by FLARE for the purpose of human examination and inspection. Currently, FLARE is non-interactive, that is, it cannot query a user for the values of missing attributes that may help improve the accuracy of its result.

#### 2.3.3 Rule Application

Let $val(a, x)$ denote the value of attribute $a$ in vector $x$. In the state of knowledge represented by a vector $v$, a rule may be applied if it *covers* $v$. A vector $x$ is said to cover a vector $y$ if and only if:

1. $x$ and $y$ have the same target-attribute and

2. for all remaining attributes $a$ of $x$, either $val(a, x) = \star$ or $val(a, x) = val(a, y)$.

For example, in $KB$, $v_{11}$ covers $v_7$ and $v_8$ but $v_{11}$ does not cover $v_9$ or $v_{12}$. Ignoring attributes whose value is $\star$, the second condition states that the set of remaining attribute-value pairs of $x$ is a proper subset of the set of remaining attribute-value pairs of $y$. Intuitively, $x$ covers $y$ if $y$ satisfies all of the premises of $x$.

To accommodate real-valued attributes, the notion of equality is slightly extended. Given that the probability of two real values being equal is extremely small, the cover relation, because of condition 2, would essentially never hold. The following extension, borrowed from ILA (Giraud-Carrier & Martinez, 1995), is suggested. Two linear values $x_1$ and $x_2$ are *equal* if and only if $|x_1 - x_2| \leq \delta$, for some $\delta > 0$. Hence, the vector ($\star$, 1.2, 3.52, $\star$, $0_T$) covers the vector (2, 1.3, 3.48, $\star$, $0_T$) if $\delta = 0.5$. In the current implementation, $\delta$ is some fraction of the range of possible values of each attribute.

#### 2.3.4 Similarity-Based Assertion

The notion of similarity in FLARE is captured by a non-symmetric distance function defined over ($n$-dimensional) vectors. If vector $x$ is stored in the knowledge base and vector $y$ is presented to the system to reason about, then the distance from $x$ to $y$ is given by:

$$D(x,y) = \frac{\sum_{i=1}^{n} d(x_i, y_i)}{num_asserted(x)}$$

where, if $x_i^+, y_i^+$ denote values of attribute $i$ other than $\star$ and ?,

$$\begin{array}{lcll} d(\star, y_i) & = & 0 & \\ d(?, y_i) & = & 0.5 & \\ d(x_i^+, ?) & = & 0.5 & \\ d(x_i^+, \star) & = & 0.5 & \\ d(x_i^+, y_i^+) & = & (x_i^+ \neq y_i^+) & \text{if attribute } i \text{ is nominal} \\ d(x_i^+, y_i^+) & = & \dfrac{|x_i^+ - y_i^+|}{range(i)} & \text{if attribute } i \text{ is linear} \end{array}$$

such that $range(i)$ is the range of values of attribute $i$ and $num_asserted(x)$ is the number of attributes that are not $\star$ in $x$. The above equations for $d$ are consistent with the semantics of $\star$ and ? defined in Section 2.1.

$D(x, y)$ is meaningful only if $x$ and $y$ have the same target-attribute and the target-attribute is left out of the computation. For example, $D(v_{11}, v_8) = 0$, $D(v_{13}, v_{14}) = 1/4$, $D(v_7, v_{16}) = 2/3$, $D(v_{16}, v_7) = 5/8$ and $D(v_1, v_4)$ is undefined. A detailed discussion of and justification for the definition of $D$ are found elsewhere (Giraud-Carrier & Martinez, 1994a). Since every ordered set is in one-to-one correspondence with a subset of the natural numbers, $D$ is well defined. To eliminate the effects of statistical outliers on $range(i)$, the dataset must be ridden of vectors whose attributes have such irregular values. $D$ is an extension of the similarity function defined for IBL (Aha, Kibler, & Albert, 1991), to inductive learning algorithms that use and/or create general rules. $D$ applies to both nominal and linear domains, and relies on the corresponding notion of distance between values. In particular, $D$ handles continuous values directly, without need for discretization. Currently, $D$ treats each attribute equally. Existing methods assigning weights to each attribute-wise distance (Salzberg, 1991; Stanfill & Waltz, 1986; Wettschereck & Dietterich, 1994) may be incorporated in $D$.

Similarity-based assertion consists of asserting the target-attribute of a vector $v$ to the value of that attribute in $v$'s closest match given by $D$. Note that (since $D$ is not symmetric) $x$ covers $y$ if and only if $D(x, y) = 0$. Hence, since 0 is the minimum of the distance function, $D$ can be used to apply both reasoning mechanisms with the correct order (i.e., rules first, similarity next), by computing the distance from all the rules in the current knowledge base to $v$ and simply selecting the rule that minimizes $D$. As it is possible that more than one rule minimizes $D$, a priority scheme is devised to choose a winner. This conflict resolution procedure which relies partially on FLARE's ability to learn is outlined in Section 2.3.5.

#### 2.3.5 Conflict Resolution

Along with each vector, FLARE also stores the following information:

- a static priority value (*static_priority*),

- a dynamic priority value (*dynamic_priority*) and
- the number of vectors covered (*num_covers*).

The value of static_priority is set to 0 by default but may be changed by a teacher to any other value a priori (e.g., $v_{11}$ and $v_{12}$ in $KB$). Static priorities provide a means whereby rules may be prioritized according to some externally provided information or meta-knowledge.

The value of dynamic_priority is initialized to 0. Its value is not changed by a teacher, however, but evolves over time and is intended to resolve conflicting defaults extensionally. Conflicting defaults, such as the Nixon Diamond of Table 1 may be encoded a priori as precepts or induced from examples. In either case, they are identified in the reasoning phase as FLARE discovers two rules that apply equally well to the input vector. Formally, two rules $R$ and $S$ are in conflict over a vector $v$ if all of the following conditions hold.

1. $D(R, v) = D(S, v) = 0$
2. $R$ and $S$ have the same specificity
3. $R$ and $S$ have the same static priority
4. $R$ and $S$ have different target-attribute's value
5. $R$ and $S$ overlap (i.e., the sets of all possible vectors each of them covers intersect)

Two vectors are said to be *concordant* if they have the same target-attribute and the target-attribute's value is the same in both vectors. When reasoning about vector $v$ and coming upon conflicting defaults, FLARE simply increments by 1 the value of dynamic_priority of the default that is concordant with $v$. If none of the defaults are concordant with $v$, then no change is made. The value of dynamic_priority reflects the number of times a particular default has been supported by evidence drawn from the environment. It is thus evidence, rather than meta-knowledge, that is responsible for the emerging ordering of defaults under dynamic_priority. Note that a target value must be given for the target-attribute for the above update to take place so that dynamic_priority is a result of the combination of learning and reasoning. Notice also that dynamic_priority evolves over time so that the system's response changes based on accumulated evidence.

The value of num_covers is also a result of the combination of learning and reasoning. It records the number of other vectors seen by the system so far that are concordant with and covered by the vector. It is a kind of confidence level for that vector, as it essentially counts the number of times the rule represented by the vector is confirmed by empirical evidence.

When more than one rule minimizes $D$ (i.e., can be selected for application), a winner is chosen according to the following priority scheme, where each subsequent condition is invoked if a tie exists at the previous level.

1. Most specific
2. Highest static priority
3. Highest dynamic priority

4. Greatest cover

Specificity is defined as the number of attributes, other than the target-attribute, whose value is not $\star$. A vector $x$ is more specific than a vector $y$ if specificity($x$)> specificity($y$). For example, in $KB$, $v_7$ has specificity 3, $v_8$ has specificity 4 and $v_{13}$ has specificity 4, so that $v_8$ and $v_{13}$ are more specific than $v_7$.

Giving priority first to the most specific vector allows FLARE to handle exceptions and cancellation of inheritance. Using static priorities next makes it possible to handle conflicting defaults as defined by a teacher, while dynamic priorities account for epistemological inconsistencies that may be resolved over time as more information becomes available in support of one belief or the other. Finally, selecting the vector with greatest cover allows evidence gathered from experience to guide a final selection. Note that the current scheme gives precedence to teacher-provided information. Other ordering schemes can easily be defined. For example, static priorities could be given as a simple form of initial bias and evidence gathered through learning (e.g., dynamic priority and cover) could be used to confirm or modify these priorities.

#### 2.3.6 Illustration

Consider $KB$ and assume the vector $v = (1, ?, 1, ?, 0, 0, ?_T)$ is input to the reasoning function. Execution proceeds as follows. Step (1) is essentially skipped as none of the attributes of $v$ meet the looping condition. Then, there are only two loops through the forward chain before the target-attribute is set.

Execution Trace
- (1) (a) Let $w = v$
  - (b) (i) Designate the second attribute as a first subgoal
    - (ii) Apply rule $v_2$: result is $w = (1, 1, 1, ?, 0, 0, ?_T)$
    - (i) Designate the fourth attribute as second subgoal
    - (ii) Apply rule $v_5$: result is $w = (1, 1, 1, 0, 0, 0, ?_T)$
  - (c) Let $v = w$
- (2) Two conflicting rules exist: $D(v_7, v) = D(v_{11}, v) = 0$
  - Apply $v_7$ (more specific): result is $v = (1, 1, 1, 0, 0, 0, 2_T)$

#### 2.3.7 Approximate Reasoning

Notice that, in forward chaining, the assertion of attributes that are subgoals does not involve similarity-based assertion but results from rule application only. As a result, the accuracy of the final goal is increased but the ability to perform approximate reasoning is reduced. It is possible to relax this restriction thus potentially achieving more subgoals but reducing the confidence in the final result. For example, the condition in step (2)(a)(ii) could be modified to allow not only rules (which are perfect matches, i.e., $D = 0$) but also matches deemed to be "close enough." The measure of closeness can be implemented via a threshold value $T_D$, placed on $D$. That is, the current condition is replaced with:

- Let $D' =$ distance to closest match
- If $D' = 0$ perform rule application
  Else if $D' \leq T_D$ perform similarity-based assertion

The value of $T_D$ then offers a simple mechanism to increase the level of approximate reasoning. This is particularly useful for cases such as the Chaco example (Section 2.3.1), where, after completion, most of the reasoning is based on the amount of similarity between concepts. Notice that the above condition is functionally equivalent to the current one when $T_D = 0$.

### 2.4 FLARE's Learning

This section addresses the construction of FLARE's knowledge base through incremental, supervised learning. FLARE learns by continually adapting to the information it receives. Indeed, training vectors are assumed to become available one at a time, over time and, as is inherent in nature, some vectors may be noisy while others may be encountered more than once. Moreover, FLARE extends inductive learning from examples with prior knowledge in the form of precepts. Sections 2.4.1 to 2.4.3 describe and illustrate FLARE's learning mechanisms and Section 2.4.4 highlights some of the advantages of combining extension and intension in learning.

#### 2.4.1 FLARE's Adapting Function

Over time, FLARE is presented with a sequence of examples and precepts that are used to update its current knowledge base. The set of all examples, rules and precepts that share the same target-attribute can be viewed as a partial function mapping instances into the goal-space. In this context, an example maps a single instance to a value in the goal-space, while precepts and rules are hyperplanes that map all of their points or corresponding instances to the same value in the goal-space.

Learning then follows a form of nearest-hyperplane learning. As mentioned in Section 2.2, it consists of first applying the reasoning scheme, and then making further adjustments to the current knowledge base to reflect the newly acquired information. The reason the algorithm is said to be nearest-hyperplane is that the reasoning phase essentially identifies a closest match for the input vector. This closest match is either a rule (i.e., true hyperplane) or a stored example (i.e., point or degenerated hyperplane).

The prior application of reasoning allows the system to predict the value of the target-attribute based on information in the current knowledge base. Also, if there are missing attributes in the input vector and the knowledge base contains rules that can be applied to assert these attributes, the rules are applied so that as many of the missing attributes as possible are asserted before the final goal is predicted. Hence, the accuracy of the prediction is increased and generalization is potentially enhanced, thus enabling FLARE to more effectively adapt its knowledge base.

The system starts with an empty knowledge base. It then adapts to each new vector $v$, where $v$ is either a precept or an example. If $v$ is the first vector, then there is no closest match and $v$ is automatically stored in the current knowledge base. In that sense, the first learned vector represents yet another bias for the learning system. If $v$ is not the first vector, then reasoning takes place producing $v^+$. A closest match, say $m$, is also found and the knowledge base adapts itself based on the relationship between $v^+$ and $m$, as shown in Figure 4. Note that $m$ is closest given the current available information and can, indeed, be "far" from $v^+$. Hence, the order in which training takes place impacts the outcome.

**DEFINITION**

- Input: the current knowledge base, the vector $v^+$ output by function Reasoning and the target value of the target-attribute.
- Output: updated knowledge base.

**IMPLEMENTATION**

1. Let $m$ be the vector of the current knowledge base such that $D(m, v^+)$ is smallest (i.e., $m$ is $v^+$'s closest match in the current knowledge base).

2. If all the attributes have equal values in $v^+$ and $m$, then add 1 to $m$.counters[$v^+$.target-attribute's value]
   *(* if m is identical to or the prototype of $v^+$, then: do not store $v^+$, update m's counters *)*

3. Else if $m$ covers $v^+$ and $m$ and $v^+$ are concordant, then add 1 to $m$.num_covers
   *(* if m subsumes or is a generalization of $v^+$, then: do not store $v^+$, increase m's confidence *)*

4. Else if $v^+$ covers $m$ and $m$ and $v^+$ are concordant, then add 1 to $v^+$.num_covers, delete $m$ from the knowledge base and add $v$ to the knowledge base
   *(* if $v^+$ subsumes or is a generalization of m, then: replace m by $v^+$, increase $v^+$'s confidence *)*

5. Else if $v^+$ and $m$ can produce a generalization, then
   *(* if there is a possibility of generalization *)*

   - If $v^+$ is more specific than $m$ and $m$ has more than one non $\star$ attribute, then drop the condition in $m$ and set $m$.static_priority to max{$m$.static_priority, $v^+$.static_priority}
     *(* if m is more general than $v^+$ and dropping condition is possible, then: do not store $v^+$, drop condition in m, update static_priority *)*

   - Else if $v^+$ has more than one non $\star$ attribute, then drop the condition in $v^+$, set $v$.static_priority to max{$m$.static_priority, $v^+$.static_priority}, set $v^+$.num_covers to $m$.num_covers, delete $m$ from the knowledge base and add $v^+$ to the knowledge base
     *(* if $v^+$ is more general than m and dropping condition is possible, then: drop condition in $v^+$, replace m by $v^+$, update parameters *)*

   - Else add $v^+$ to the knowledge base
     *(* if dropping condition is impossible, then: store $v^+$ in knowledge base *)*

6. Else add $v^+$ to the knowledge base
   *(* default case: store $v^+$ in the knowledge base *)*

Figure 4: Function Adapting

The array *counters* contains an entry for each possible value of the target-attribute and is also stored with each vector. All the counters are initialized to 0, except the one corresponding to the vector's target-attribute value which is initialized to 1. The counters evolve over time and are used to handle noise. For any vector $p$ in the current knowledge base, exactly one counter value is incremented (by 1) each time a new vector is presented, whose attributes' values are all equal to those of $p$. The value incremented corresponds to the new vector's target-attribute value. The counter value that is highest represents the statistically "most probable" target-attribute value. In effect, the target-attribute value of a vector is always the one with highest count. Note that this value may change over time, as new information becomes available.

Because a best match is first identified, changes to the knowledge base are localized and are guided by the kinds of possible relationships between $v^+$ and $m$. These relationships are summarized below.

- $v^+$ is equal to $m$ (i.e., noise or duplicates) or
- $v^+$ is subsumed by $m$ (i.e., $v^+$ is a special case of $m$) or
- $v^+$ subsumes $m$ (i.e., $v^+$ is a general case of $m$) or
- $v^+$ and $m$ can produce a generalization or
- all other cases (e.g., $v^+$ is an exception to $m$, $v^+$ and $m$ are too far apart, etc.)

In the first case, $v^+$ (or its prototype $m$) is already in the knowledge base and only the counters need to be updated. Note that the extension of the notion of equality discussed in Section 2.3.3 enables this part of the algorithm, in conjunction with the counters, to produce some generalization for linear attributes. In effect, the vector retained in the knowledge base acts as a "prototype," and its target-attribute's value is the one most probable among its $\delta$-close neighbors. In the second case, there is no need to store $v^+$ as the current knowledge base has sufficient information to correctly predict $v^+$'s target value. In the third case, $v^+$ is stored and $m$ removed as $v^+$ is more general than $m$ and thus accounts for it. The fourth case captures the possibility of generalization by dropping conditions (see Section 2.4.2 for details). If generalization takes place, only one of $v_+$ or $m$ is generalized and stored. The values of static_priority and num_covers are also reset so that the generalization inherits the maximum static priority value and the current value of num_covers. Finally, in the fifth case, $v^+$ must be added as the current knowledge base either does not produce the correct target value for $v^+$ (e.g., exceptions) or is not deemed reliable enough to properly account for $v^+$.

Notice that the adaptation phase takes place regardless of the target value predicted by the reasoning phase. A possible alternative would be to adapt only if the predicted target value differs from the actual target value. It has been found empirically, however, that too much useful information is lost with this approach due to the incrementality of the system and its sensitivity to ordering. A possibly viable alternative would make use of memory. Vectors currently accounted for could be saved in memory and presented later to the system. This may be done a few times over some period of learning time until either the vectors must be stored in the knowledge base (due to changes in the knowledge base) or they are discarded as the system has gained enough confidence in its ability to account for them.

#### 2.4.2 Generalization

Two vectors that have the same target-attribute's value can produce a generalization when the following five conditions hold.

1. They differ in the value of exactly one of their attributes.
2. The attribute on which they differ is nominal.
3. They are concordant.
4. The number of their attributes not equal to $\star$ differ by at most 1.
5. At least one of them has more than one non $\star$ attribute.

Generalization then consists of setting to $\star$ the attribute on which the two vectors differ, in the vector that is most general, as long as that vector has more than one non $\star$ attribute. For example, vectors $v_8$ and $v_9$ of $KB$ satisfy the above conditions and would generalize to produce a vector, say $v_{8+9} = (\star, \star, \star, 0, 0, 1, 3_T)$. The value of 1 in the fourth condition is based upon empirical evidence.

Choosing the most general vector maximizes generalization and the condition on the number of non $\star$ attributes guarantees that no rule is generated that would cover every other vector. This version of the *dropping-the-condition* rule (Michalski, 1983) is only applied to nominal attributes as it makes little sense for linear (especially real-valued) domains. For linear attributes, generalization is achieved through the artifact due to the extended notion of equality discussed above.

Let $v$ and $w$ be two vectors representing $n$ and $m \leq n$ examples, respectively. Furthermore, let $v'$ be the generalization obtained from $v$ and $w$ by dropping a $p$-valued attribute in $v$. Then $v'$ represents $pn$ examples. Since $v$ and $w$ represent at most $2n$ examples, generalization causes at least $(p-2)n$ new examples to be represented. As $p$ increases, this value also increases and, for large values of $p$, could lead to over generalization as only two values of a given attribute suffice to predict the outcome of all values in the current context. However, such potential over generalizations are partially offset by the system's ability to identify, retain and give precedence to exceptions.

There are still drawbacks to FLARE's generalization scheme. Given a set of vectors of the form $v_i = Sx_iT_k$ where $S$ is fixed, $T$ is the target-attribute and $x_i \neq x_j$ for all $i \neq j$, any pair of concordant (i.e., same $k$) vectors satisfy the generalization condition, yet only the first such pair will generalize. All other vectors then become either subsumed by this generalization or exceptions to it. If most are exceptions, this leads to the storage of more vectors than needed, especially for large domains where various subsets of values give rise to different target-attribute's values. Moreover, the outcome depends upon ordering of the vectors. Also, if there exists conflicts involving one (or more) value of $x$, then the system will end up giving unfounded precedence to the exceptions (being more specific) and, again, these depend on the ordering. Support for internal disjunction or a more complex generalization scheme may help alleviate some of these problems. They are the topic of future research.

### 2.4.3 Illustration

This section shows the evolution of FLARE's knowledge base as the vectors of $KB$ (see Section 2.1.3) are presented to it as inputs. It highlights several interesting features of both reasoning and adaptation. Let $KB'$ denote the current knowledge base of FLARE. As discussed above, FLARE starts with $KB' = \emptyset$. Each vector is presented to FLARE in the order in which it appears in $KB$.

1. Presentation of $v_1$. $KB' = \emptyset$. $v_1$ is simply added to $KB'$.

2. Presentation of $v_2$. $KB' = \{v_1\}$. $v_1$ is the closest match. $v_1$ and $v_2$ are not concordant, so $v_2$ is added to $KB'$.

3. Presentation of $v_3$. $KB' = \{v_1, v_2\}$. Same as above with either $v_1$ or $v_2$. So $v_3$ is added to $KB'$.

4. Presentation of $v_4$. $KB' = \{v_1, v_2, v_3\}$. No winner can be found since none of the vectors in $KB'$ have the same target-attribute as $v_4$. So $v_4$ is added to $KB'$. While the earlier $KB'$ had information about a single concept (i.e., stimulus-situation), the new $KB'$ now provides FLARE with knowledge about a new concept, namely appropriate-response. By "partitioning" vectors along their target-attribute, FLARE naturally supports multiple concept learning.

5. Presentation of $v_5$. $KB' = \{v_1, v_2, v_3, v_4\}$. $v_4$ is the only (and hence closest) match since none of the other vectors in $KB'$ have the same target-attribute as $v_5$. $v_4$ and $v_5$ satisfy conditions 1-4 for generalization but violate condition 5, so $v_5$ is added to $KB'$.

6. Presentation of $v_6$. $KB' = \{v_1, v_2, v_3, v_4, v_5\}$. Similar to step 3 with $v_4$ and $v_5$. So $v_6$ is added to $KB'$.

7. Presentation of $v_7$. $KB' = \{v_1, v_2, v_3, v_4, v_5, v_6\}$. Note that $v_7$ is a precept. No winner can be found since none of the vectors in $KB'$ have the same target-attribute as $v_7$. So $v_7$ is added to $KB'$. A third concept, namely media, is now available.

8. Presentation of $v_8$. $KB' = \{v_1, v_2, v_3, v_4, v_5, v_6, v_7\}$. $v_7$ is the only (and hence closest) match since none of the other vectors in $KB'$ have the same target-attribute as $v_8$. $v_8$ is an exception to $v_7$ since $v_7$ covers $v_8$ but they are not concordant. Hence, $v_8$ is added to $KB'$. Though $v_7$ suggests to use lecture as a media, the added condition on training-budget found in $v_8$ causes that suggestion to change to lecture with slides.

9. Presentation of $v_9$. $KB' = \{v_1, v_2, v_3, v_4, v_5, v_6, v_7, v_8\}$. Only $v_7$ and $v_8$ may compete. $D(v_7, v_9) = 1/3$ and $D(v_8, v_9) = 1/4$. Hence, $v_8$ wins. $v_8$ and $v_9$ satisfy all five conditions for generalization. The second attribute is dropped (i.e., replaced by $\star$) in either one, say $v_8$, to produce $v_8' = (\star, \star, \star, 0, 0, 0, 3_T)$. $v_8'$ is added to $KB'$. All of the attributes in $v_8$ and $v_9$ have the same value, except for stimulus-situation. This is sufficient for FLARE to hypothesize that the value of stimulus-situation is not critical and the attribute may thus be ignored. In other words, FLARE decides that the value of stimulus-situation is not needed when predicting lecture with slides.

10. Presentation of $v_{10}$. $KB' = \{v_1, v_2, v_3, v_4, v_5, v_6, v_7, v_8'\}$. Only $v_7$ and $v_8'$ may compete. $D(v_7, v_{10}) = 1/3$ and $D(v_8', v_{10}) = 0$. Hence, $v_8'$ wins. $v_8'$ covers $v_{10}$ and they are concordant, so FLARE adds 1 to num_covers($v_8'$). $v_{10}$ need not be added to $KB'$. $v_{10}$ is one of the many special cases now handled by the new generalization $v_8'$.

11. Presentation of $v_{11}$. $KB' = \{v_1, v_2, v_3, v_4, v_5, v_6, v_7, v'_8\}$. Notice that $v_{11}$ is also a precept, so that precepts may be given at any time during learning. Only $v_7$ and $v'_8$ may compete. $D(v_7, v_{11}) = 1/6$ and $D(v'_8, v_{11}) = 1/3$. Hence, $v_7$ wins. Neither one covers the other; they are not equal; they cannot produce generalization (violate condition 3). Thus, $v_{11}$ is added to $KB'$. Note that $v_{11}$ has a static priority of 1.

12. Presentation of $v_{12}$. $KB' = \{v_1, v_2, v_3, v_4, v_5, v_6, v_7, v'_8, v_{11}\}$. $v_7$, $v'_8$ and $v_{11}$ compete. $D(v_7, v_{12}) = 1/2$, $D(v'_8, v_{12}) = 2/3$ and $D(v_{11}, v_{12}) = 1/4$. Hence, $v_{11}$ wins. Neither one covers the other; they are not equal; they cannot produce generalization (violate condition 3). Thus, $v_{12}$ is added to $KB'$. Note that $v_{12}$ has a static priority of 3. Since $v_{11}$ and $v_{12}$ overlap, precedence would be given to $v_{12}$ in case of a conflict.

13. Presentation of $v_{13}$. $KB' = \{v_1, v_2, v_3, v_4, v_5, v_6, v_7, v'_8, v_{11}, v_{12}\}$. In this case, some non-asserted attributes of $v_{13}$ may be asserted through reasoning, before the target-attribute. Rules $v_2$ and $v_4$ are applied to assert the second and fourth attributes respectively. The result is $v'_{13} = (1, 1, 0, 0, 0, 1, 2_T)$. $v_7$, $v'_8$, $v_{11}$ and $v_{12}$ compete to assert the target-attribute. $D(v_7, v'_{13}) = 0$, $D(v'_8, v'_{13}) = 0$, $D(v_{11}, v'_{13}) = 0$, and $D(v_{12}, v'_{13}) = 1/2$. Both $v_7$ and $v'_8$ win over $v_{11}$ since they are more specific. However, $v_7$ and $v'_8$ have the same specificity. In fact, they satisfy all five conditions that identify them as conflicting defaults in the current context. Hence, FLARE adds 1 to dynamic_priority($v_7$) since $v_7$ and $v'_{13}$ are concordant. Reasoning then proceeds, giving precedence to $v_7$. Since $v_7$ covers $v'_{13}$ and they are concordant, $v'_{13}$ need not be added to $KB'$.

14. Presentation of $v_{14}$. $KB' = \{v_1, v_2, v_3, v_4, v_5, v_6, v_7, v'_8, v_{11}, v_{12}\}$. Some non-asserted attributes of $v_{14}$ may be asserted through reasoning, before the target-attribute. Rules $v_2$ and $v_5$ are applied to assert the second and fourth attributes respectively. The result is $v'_{14} = (1, 1, 1, 0, 0, 1, 2_T)$. The rest is identical to step 13. Now, dynamic_priority($v_7$) = 2 and $v'_{14}$ is not added to $KB'$.

15. Presentation of $v_{15}$. $KB' = \{v_1, v_2, v_3, v_4, v_5, v_6, v_7, v'_8, v_{11}, v_{12}\}$. Some non-asserted attributes of $v_{15}$ may be asserted through reasoning, before the target-attribute. Rules $v_3$ and $v_6$ are applied to assert the second and fourth attributes respectively. The result is $v'_{15} = (2, 2, 2, 1, 1, 0, 0_T)$. $v_7$, $v'_8$, $v_{11}$ and $v_{12}$ compete to assert the target-attribute. $D(v_7, v'_{15}) = 1$, $D(v'_8, v'_{15}) = 1$, $D(v_{11}, v'_{15}) = 1$, and $D(v_{12}, v'_{15}) = 1/2$. Hence, $v_{12}$ wins. Neither one covers the other; they are not equal; they cannot produce generalization (violate condition 3). Thus, $v'_{15}$ is added to $KB'$.

16. Presentation of $v_{16}$ and $v_{17}$. $KB' = \{v_1, v_2, v_3, v_4, v_5, v_6, v_7, v'_8, v_{11}, v_{12}, v'_{15}\}$. Both are equal to $v'_{15}$. Neither $v_{16}$ nor $v_{17}$ need be added to $KB'$ but the appropriate counter values are incremented in $v'_{15}$. The result is counters[0] = 2, counters[1] = 0, counters[2] = 1 and counters[3] = 0. Thus, the target-attribute's value of $v'_{15}$ is currently 0.

The resulting $KB'$, after processing $KB$ is shown in Figure 5. The variables $p$, $c$ and $dp$ stand for static priority, cover number and dynamic priority, respectively. At an intuitive level, FLARE has used both learning and reasoning mechanisms to deal with $KB$. Induction (on vectors $v_8, v_9, v_{10}$) has allowed the system to decide that the stimulus-situation was irrelevant in predicting the use of lecture-with-slides. Deduction from the empirical evidence provided by vectors $v_{13}$ and $v_{14}$ has caused FLARE to break the "tie" between rules $v_7$ and $v'_8$ in favor of $v_7$. Prior knowledge relative to vectors $v_{11}$ and $v_{12}$ was encoded as static priorities, thus giving precedence to $v_{12}$ in case of conflicts. Hence, if the vector $v = (1, ?, 0.?, 0, 1, ?_T)$ is presented to FLARE after $KB'$ is acquired, the second and fourth attributes are first asserted as previously discussed to produce $v' = (1, 1, 0, 0, 0, 1, ?_T)$. Then, $v_7$, $v'_8$ and $v_{11}$ compete. $v_7$ and $v'_8$ win due to specificity. $v_7$ and $v'_8$ also have same static priorities but $v_7$ wins due to dynamic priority and the result is $(1, 1, 0, 0, 0, 1, 2_T)$. Now, if the vector $(1, ?, 2, ?, 0, 0, ?_T)$ is presented, a similar situation arises between $v_{11}$ and $v_{12}$. The conflict is resolved with static priorities.

| | | | | | | | | | |
|---|---|---|---|---|---|---|---|---|---|
| $v_1$ | = | 0 | $0_T$ | $\star$ | $\star$ | $\star$ | $\star$ | $\star$ | $(p = c = dp = 0)$ |
| $v_2$ | = | 1 | $1_T$ | $\star$ | $\star$ | $\star$ | $\star$ | $\star$ | $(p = c = dp = 0)$ |
| $v_3$ | = | 2 | $2_T$ | $\star$ | $\star$ | $\star$ | $\star$ | $\star$ | $(p = c = dp = 0)$ |
| $v_4$ | = | $\star$ | $\star$ | 0 | $0_T$ | $\star$ | $\star$ | $\star$ | $(p = c = dp = 0)$ |
| $v_5$ | = | $\star$ | $\star$ | 1 | $0_T$ | $\star$ | $\star$ | $\star$ | $(p = c = dp = 0)$ |
| $v_6$ | = | $\star$ | $\star$ | 2 | $1_T$ | $\star$ | $\star$ | $\star$ | $(p = c = dp = 0)$ |
| $v_7$ | = | $\star$ | 1 | $\star$ | 0 | 0 | $\star$ | $2_T$ | $(p = c = 0, dp = 2)$ |
| $v'_8$ | = | $\star$ | $\star$ | $\star$ | 0 | 0 | 1 | $3_T$ | $(p = 0, c = 1, dp = 0)$ |
| $v_{11}$ | = | $\star$ | 1 | $\star$ | $\star$ | 0 | $\star$ | $0_T$ | $(p = 1, c = dp = 0)$ |
| $v_{12}$ | = | $\star$ | 1 | $\star$ | 1 | $\star$ | $\star$ | $1_T$ | $(p = 3, c = dp = 0)$ |
| $v'_{15}$ | = | 2 | 2 | 2 | 1 | 1 | 0 | $0_T$ | $(p = c = dp = 0)$ |

Figure 5: $KB'$

### 2.4.4 Extensionality and Intensionality

As it is able to use prior knowledge in the form of precepts together with raw examples, FLARE effectively combines the intensional approach (based on features, expressed here by precepts) and the extensional approach (based on instances, expressed by examples) to learning and reasoning. With this combination, FLARE can resolve conflicting defaults, such as the Nixon Diamond (Reiter & Griscuolo, 1981), by either being told explicitly which default prevails (e.g., religious conviction is more important than political affiliation) or by computing relative dynamic priorities (see Section 2.3.5) from examples of Republican-Quakers.

Most inductive learning systems are purely extensional, while most reasoning systems are purely intensional. It is therefore these authors' contention that, if induction and deduction are to be integrated, then a combination of the two approaches is desirable. It is also clear that the combination increases flexibility. On the one hand, extensionality accounts for the system's ability to adapt to its current environment, i.e., to be more autonomous. On the other hand, intensionality provides a mechanism by which the system can be taught and thus does not have to unnecessarily suffer from poor or atypical learning environments.

In the context of reasoning, precepts provide a useful medium to encode certain first-order language statements (e.g., the rule base of an expert system) that can, in turn, be learned by FLARE (in the usual way) and later be used for reasoning purposes.

**DEFINITION**

- Input: a set of general rules, a set of facts and one designated target-attribute
- Output: one or more precepts

**IMPLEMENTATION**

1. **Learning general knowledge:** Perform Learning on the set of general rules.
2. **Reason from facts:** Perform Reasoning with a vector encoding the given facts and the designated target-attribute.

Figure 6: Function Generate-Precepts

## 2.5 FLARE's Automatic Generation of Precepts

Section 2.1.2 introduced the notion of precepts as generalized AVL vectors in which some of the attributes have the special value $\star$ (i.e., don't-care). Precepts may be encoded directly by a teacher or deduced automatically from general knowledge. FLARE provides a simple (off-line) mechanism for the automatic generation of precepts in the preprocessing phase described in Section 2.2.

FLARE uses prior knowledge in the form of general rules that may be viewed as encoding "commonsense" knowledge involving some of the attributes of the application domain. With the appropriate setting for deduction, FLARE can then generate domain-specific precepts that can be used as biases for inductive learning or for further reasoning about the specific domain to which they apply.

Consider the example in Table 3 from Section 2.1.1. Assume that the system is to inductively learn rules regarding the suitability of lenses for patients from a set of examples whose attributes include the patient's tear-production rate (tpr). The statements in Table 3 capture general knowledge about eyes. Informally, they state that:

1. Low tear-production rate causes dryness of the eyes.

2. Dry eyes are not fit for lenses.

When provided with the fact that the target-attribute of the system has to do with fitting lenses, the general knowledge may be used to produce a domain-dependent precept that states that, if a patient has low tear-production rate, then he/she should not be fitted lenses. The precept, in turn, provides a useful bias to the system during further induction from examples.

The process of generating precepts described above is essentially one of acquiring the general knowledge (or rules) and reasoning from it as described in Figure 6. When general rules are available, the function Generate-Precepts is always invoked prior to any other work by FLARE.

The function Generate-Precepts actually makes use of the other functions of FLARE. In step (1), it constructs a knowledge base from the general rules using learning as described in Section 2.4. In step (2), it reasons, as described in Section 2.3, using the acquired knowledge and facts enabling the general knowledge to be applied to the domain. The facts are encoded as a vector in which attributes found in the general knowledge are set to appropriate values and all others set to $\star$. Since precepts are mostly used as learning biases, the designated target-attribute is typically the target concept of an inductive application. In the lenses example of Table 3, the appropriate setting is obtained by creating a vector such that attribute $Tpr$ is set to low and attribute $Fit$ is designated as the target-attribute. Having incorporated the two rules in its knowledge base in step (1), FLARE would then easily deduce a precept of the form: if $Tpr$ is low then $Fit$ is false, independent of any other don't-care conditions.

Though the function Generate-Precepts is automated, the setting of the relevant attributes and interpretation of the result rely on a teacher. More automatic mechanisms may be considered, where the system could try any combination of a learning problem's attributes values to instantiate general knowledge. Then, any such instantiation that causes the target-attribute to become asserted is a potential precept. However, that process would be exponential and most of it would probably not lead to any useful conclusion.

## 3. Experimental Results and Demonstrations

A set of classical commonsense benchmark problems has been proposed by Lifschitz (1988) and the UCI repository (Murphy & Aha, 1992) contains many useful training sets for inductive learning. This section reports results obtained with FLARE on several of these datasets. Results on a number of other uses of the framework, including two expert systems, are also presented. Finally, some of the limitations of the system are described.

One artifact of the implementation is that, since variables cannot be added dynamically, all attributes must be defined a priori. All attributes that do not appear in rules, examples or precepts are set to don't-care. This is consistent with the semantics of don't-care and does not interfere with the algorithm since the distance $D$ essentially treats learned don't-cares as neutral values.

### 3.1 Inductive Learning and Prior Knowledge

In order to test the predictive accuracy of FLARE, the standard training set/test set approach is used. The value of $v$'s target-attribute is provided but it is not used during reasoning. Rather, the system reasons based on its current knowledge base and all of the asserted attributes of $v$. When reasoning is completed, the "computed" target value is compared with the "actual" target value.

Several datasets from the UCI repository (Murphy & Aha, 1992) were chosen. They represent a wide variety of situations, as shown in Table 6. The column labelled "Size" indicates the total number of examples in the dataset. The column labelled "Attributes" records the number and type (L for linear, N for nominal) of all the attributes, other than the target (or output) attribute. The column labelled "Output" shows the number of output classes.

FLARE's results were gathered for each of the above applications, using 10-way cross-validation. Each dataset is randomly broken into 10 sets of approximately equal size. Then, in each *turn*, one of the sets is used for testing, while the remaining 9 are used for learning. This process is repeated 10 times, one for each test set, so that every item of data is in the test set once and only once. Because FLARE's outcome is dependent upon the ordering of data during learning, each turn was repeated 10 times with a new random ordering of the training set. The predictive accuracy for a given turn is the average of the 10 corresponding trials and the predictive accuracy for the dataset is the average of the 10 turns.

| Application | Size | Attributes | Output |
|---|---|---|---|
| lenses | 24 | 4N | 3 |
| voting-84 | 435 | 16N | 2 |
| tic-tac-toe | 958 | 9N | 2 |
| hepatitis | 155 | 6L&13N | 2 |
| zoo | 90 | 16N | 7 |
| iris | 150 | 4L | 3 |
| soybean-small | 47 | 4L&31N | 4 |
| segmentation | 420 | 19L | 7 |
| glass | 214 | 9L | 7 |
| breast-cancer | 699 | 9L | 2 |
| sonar | 208 | 60L | 2 |

Table 6: Selected Applications

| Application | PA | IR | ID3 | CN2 | BP |
|---|---|---|---|---|---|
| lenses | 79.0 | .43 | 65.0 | 83.3 | 76.7 |
| voting-84 | 92.9 | .63 | 95.4 | 93.8 | 96.0 |
| tic-tac-toe | 81.5 | 1.0 | 85.6 | 98.0 | 96.6 |
| hepatitis | 80.0 | .94 | 77.9 | 76.1 | - |
| zoo | 97.4 | .36 | 97.8 | 93.3 | 97.8 |
| iris | 94.0 | .13 | 94.0 | 93.3 | 97.3 |
| soybean-small | 100 | .98 | 98.0 | 100 | 100 |
| segmentation | 94.0 | .99 | 96.9 | 94.1 | - |
| glass | 71.8 | .22 | 67.7 | 62.7 | 38.0 |
| breast-cancer | 96.6 | .47 | 95.1 | 95.1 | 99.7 |
| sonar | 83.8 | .77 | 77.4 | 44.3 | - |
| Averages | 88.3 | .63 | 86.4 | 84.9 | 87.8 |

Table 7: FLARE: Induction

Results are shown in Table 7. The first number (PA) represents predictive accuracy (in %) on the test set after training and the second number (IR) is the inductive ratio, defined as the ratio of the size (in number of rules) of the final knowledge base to the number of instances used in learning. IR is another measure of the generalization power of FLARE, as well as an indication of FLARE's memory requirements. Results of PA with ID3 (Quinlan, 1986), ordered CN2 (Clark & Niblett, 1989) and Backpropagation (Rumelhart & McClelland, 1986) are also included for comparison. They were also obtained using 10-way cross-validation and are as reported by Zarndt (1995).

For the set of selected applications, FLARE's performance in generalization compares favorably with that of ID3, CN2 and Backpropagation, as well as with that of other inductive learning algorithms (e.g., see Aha et al., 1991; Wettschereck & Dietterich, 1994; Zarndt, 1995). In addition, the knowledge base maintained by FLARE is generally significantly smaller than the set of all training vectors.

| Application | no prec. | w/prec. |
|---|---|---|
| lenses | 79.0 - .43 | 80.5 - .33 |
| voting-84 | 92.9 - .63 | 94.5 - .25 |
| tic-tac-toe | 81.5 - 1.0 | 88.5 - .72 |
| hepatitis | 80.0 - .94 | 81.2 - .68 |
| zoo | 97.4 - .36 | 97.4 - .32 |
| Averages | 86.2 - .67 | 88.4 - .46 |

Table 8: FLARE: Induction with Prior Knowledge

The first five applications were further selected to illustrate the effect of prior knowledge on predictive accuracy and inductive ratio. For each application, the above experimental procedure is repeated but the set of training examples is now augmented by precepts given a priori (i.e., before the training set is presented). Results are reported in Table 8. Each column shows both PA and IR. Here, the precepts are obtained from domain knowledge provided with the application (voting-84) or generated from the authors common sense (zoo, lenses, hepatitis, tic-tac-toe). They serve as learning biases. The results with precepts show an average increase of 2.6% in predictive accuracy and a decrease of 31.3% of the inductive ratio. The decrease in IR demonstrates that prior knowledge allows pruning of parts of the input space during learning. Indeed, starting with the same number of training vectors, FLARE ends up with a knowledge base containing about one-third less vectors than when precepts are not used. Hence, precepts not only increase generalization performance, they also reduce memory requirements.

The *lenses* application was also used to demonstrate how precepts may be generated automatically by deducing domain-dependent information from general knowledge, as discussed in Section 2.5. The example of Table 3 from Section 2.1.1 was implemented (as described in Section 2.5) and a precept stating that, if the Tpr attribute is set to low, then lenses should not be prescribed was generated. That precept was, in turn, used prior to performing inductive learning as described above.

The process of inductive learning with automatically generated prior knowledge is two-phase, where both phases perform the same operations on different pieces of information. In the first phase, general knowledge expressed as rules (and translated into AVL) is learned by FLARE. Then FLARE reasons based on some instantiation that links the general knowledge to the current domain. The result of this reasoning phase is one (or more) precept containing domain-dependent information. In the second phase, FLARE learns from the generated precepts and any other available examples. The result is a set of inductively generated rules.

### 3.2 Classical Reasoning Protocols

Several problems from the set of Benchmark Problems for Formal Nonmonotonic Reasoning (Lifschitz, 1988), were presented to FLARE. The problems were first translated into their corresponding AVL representation. FLARE is able to properly incorporate the premises and correctly derive the expected conclusions for the following classes of problems from (Lifschitz, 1988):

- A1 - basic default reasoning.
- A2 - default reasoning with irrelevant information
- A3 - default reasoning with several defaults
- A5 - default reasoning in an open domain
- A9 - priority between defaults
- B1 - linear inheritance (top-down)
- B2 - tree-structured inheritance
- B3 - one-step multiple inheritance
- B4 - multiple inheritance

Problem A4 involves a disabled default and problems A6 through A8 deal with unknown exceptions. Such problems cannot be represented in FLARE. Problems A10 and A11 deal with instances of defaults and reasoning about priority. Though not directly representable in FLARE, they are effectively solved via the use of static (A10) or dynamic (A11) priorities. The other classes of problems defined by Lifschitz (1988) (i.e., reasoning about actions, uniqueness of names and autoepistemic reasoning) are beyond the current scope of FLARE.

Note that, in order to work properly, some of the above problems require added processing. In particular, problems A1, A2, A3 and A5 involve both classes of objects and particular instances of these classes. Problem A1, for example, is given as follows: blocks A and B are heavy, heavy blocks are normally located on the table, A is not on the table. Translating to AVL gives: A $1_T$ $\star$, B $1_T$ $\star$, $\star$ 1 $1_T$ and A $\star$ $0_T$, where the first attribute is a multi-valued attribute representing the objects in the universe and the second and third attributes are Boolean, encoding the predicate *heavy* and $on_table$ respectively. Now, if A ? $?_T$ is shown, A 1 $?_T$ will be derived from the first vector and will then match both $\star$ 1 $1_T$ and A $\star$ $0_T$ exactly. It seems reasonable that priority should be given to the later since it involves A (an instance) explicitly. To solve this problem, vectors involving explicit references to instances of objects have their static priority set to 1 while all other vectors have their static priority set to 0. This is, of course, an artifact of encoding. An alternative is to write all facts relative to a given instance as a definition whose target-attribute is the instance value. Then completion would guarantee the correct outcome.

The above problems are characteristics of important forms of human patterns of reasoning. However, they are artificial as they have been manufactured explicitly with the intent of isolating one salient feature of nonmonotonic reasoning, independent of all others. To further investigate the properties of FLARE and the combination of learning and reasoning, other more "real-world" applications must be designed and experimented with. Section 3.5 presents such preliminary applications. The next two sections present simple applications that further exercise FLARE's ability to learn incrementally and to combine learning with reasoning in useful ways.

### 3.3 The Nixon Diamond

The Nixon Diamond (Reiter & Griscuolo, 1981), reproduced as Table 1 in Section 2.1.1, is important as a prototype of a class of interesting problems involving conflicting defaults. It is used here to demonstrate FLARE's mechanisms to handle such conflicts both intensionally and extensionally.

FLARE's static priorities offer a simple way of resolving the Nixon Diamond intensionally, based on some externally provided information (e.g., religious convictions supersede political affiliations). In that case, both defaults are given along with an appropriate static priority.

Another alternative consists of providing the defaults without any priority. This corresponds to a possibly more natural situation where the system really is in a don't-know state when it comes to deciding on Nixon's dispositions. Yet, such don't-know states are uncomfortable and it is the authors' contention that any kind of information that may allow a decision to be made should be used. Hence, a simple epistemological approach is adopted, where the conflict arises due to beliefs rather than facts. In this case, it is possible to attempt to resolve the conflict by observing instances of Republican-Quaker. The relative number of pacifists and non-pacifists can then serve as evidence to lean towards one decision or the other. In other words, it is the system's observation of what seems most common in its environment that creates its belief. This is not unlike the way humans deal with many similar situations.

A final approach, which combines inductive learning and reasoning, consists of not providing the system with any default. Rather, examples of Republicans, Quakers and Republican-Quakers are shown and the system automatically comes up with both the defaults (through induction) and their relative priorities.

All three of these experiments were run with FLARE and the results are as expected. In the third case, the actual knowledge base depends upon the ordering. It consists of one vector for Republicans that are not Quakers or one vector for Quakers that are not Republicans, one default vector for Quakers or Republicans and one vector for Republican-Quakers. The target value of the vector for Republican-Quakers is not had via dynamic_priority but via the counters. Functionally, however, the result is identical.

### 3.4 Do Birds Typically Fly?

Incremental learning is one of FLARE's important features. With incrementality, the system is self-adaptive in the sense that its current knowledge base is representative of its experience with its environment so far. And the knowledge base can be continually updated as new information becomes available. To exercise incrementality a simple example of bottom-up inheritance involving birds was designed.

The application has four attributes, two of which correspond to Ostrich and Bird. The other two are other (undetermined) attributes of birds (e.g., Feather). The target-attribute is Boolean and characterizes the ability to fly. At first, the system is exposed mostly to ostrich-birds (maybe the experiment is started in Australia). When asked whether birds typically fly (i.e., only the Bird attribute is asserted and all other inputs are don't-know), FLARE concludes that birds do not fly, which is consistent with its current experience with the "world." However, as more new instances of flying birds (i.e., other than ostriches and penguins) are encountered, FLARE adapts its knowledge and when asked again, concludes that birds fly. Correct knowledge about ostrich-birds is also preserved. That is, if the system is shown an ostrich, it will still conclude that the ostrich does not fly.

Of course, a precept may also be given to the system at any given time, stating that birds typically fly. The idea is that FLARE offers both options naturally. The system may be taught so as not to suffer from poor or atypical learning environments (e.g., Australia for birds' flying ability prediction), or it may be left to adapt to its environment. As research on autonomous agents continues, this later ability becomes important.

Note that the above example also illustrates one of FLARE's limitations. The system either concludes that birds do fly or that they do not. There is no mechanism for representing a middle ground in such a way that FLARE could reason about it at the meta-level. Decisions made in the presence of conflicts are also "crisp" as demonstrated by the simple, rigid conflict resolution mechanism discussed in Section 2.3.5. Even though, the system may be able to produce more fuzzy-like results by associating each decision with a confidence level, it would still not be able to reason about these at the meta-level.

### 3.5 Learning Expert Systems

In order to better assess FLARE's reasoning mechanisms, two expert system knowledge bases are used. One is called *mediadv* (Harmon & King, 1985) and is intended to help designers or committees choose the most appropriate media to deliver a training program. It consists of 20 rules with chains of inference of length 2 at most. The other is called *health* (Sawyer & Foster, 1986) and is intended to predict the longevity of patients based on a variety of factors (e.g., weight, personality, etc.). It is much larger as it contains 77 rules and more complex as it involves longer chains of inference. Five rules were left out as one is redundant (i.e., rule 17 is identical to rule 14) and four are only needed in the interactive setting in which the original system is described. Hence, only 72 rules are considered.

Both sets of rules were translated into AVL. The 20 rules of *mediadv* produce 99 vectors and the 72 rules of *health* produce 72 vectors. The number of vectors for *mediadv* is much larger than the original number of rules because many of the rules contain internal disjunctions. In AVL, a new vector must be constructed for each possible combination arising from the disjunctions. For example, the rule: If ((A=1 or A=2 or A=5) and (B=2.9 or B=7.8)) then C, gives rise to 6 vectors corresponding to the equivalent set of rules: If ((A=1) and (B=2.9)) then C, If ((A=1) and (B=7.8)) then C, If ((A=2) and (B=2.9)) then C, etc.

The sets of vectors corresponding to the original knowledge bases are not encoded into FLARE. Rather, they are presented to the system to be learned. Hence, some generalization may take place. In fact, the final number of vectors (after learning) in *mediadv* is only 71, while in *health* it is 65.

The *mediadv* example is clearly very simple and presents little interest in terms of deduction. However, its purpose here is to show how the system's current knowledge base may be updated through learning. Of particular interest is the case of conflicts that arise because two or more rules may apply to a given situation, while implying different goal values. In *mediadv*, such a conflict exists between rules 13 and 14 and between rules 16 and 17. Rules 13 and 14 are used as illustration. Let X be some fixed conjunction of conditions not shown. Then:

- rule 13: if (X) and (training_budget = small or training_budget = medium) then media_to_consider = lecture
- rule 14: if (X) and (training_budget = medium) then media_to_consider = lecture-with-slides

It is clear that, in some cases, these rules conflict. The important issue is that it is difficult to avoid such occurrences in large knowledge bases elicited from experts. As FLARE supports learning, it is possible, however, to look at various (historical) situations where training_budget was medium and check which media was used then. This information can, in turn, be used to give precedence to one rule over the other. Moreover, this precedence need not be fixed after so many examples have been considered. Indeed, it may evolve over time and even change radically depending on circumstances.

An example, using several additional instances of [(X) and (training_budget = medium)] together with a target value for media_to_consider, was implemented. The instances used caused the value of dynamic_priority of rule 13 to be greater than that of rule 14, thus effectively giving (evidential) precedence to rule 13.

Our experiments with the *health* knowledge base demonstrate FLARE's ability to perform deduction. The experiments conducted involve chains of inference of reasonable lengths and are fairly intuitive. Results are summarized in Table 9. The first column contains the list of attributes used in the knowledge base. Then, each pair (setting, result) of columns represents an experiment in reasoning with the knowledge base. The setting column contains the data FLARE starts with. Unknown conditions (or attributes) are initialized as don't-knows (i.e., ?). The result column shows the state of knowledge after reasoning. Each derived piece of information is italicized and subscripted by the depth of inference at which it was derived.

Starting with the facts in the setting column, FLARE successively infers new conclusions until it reaches a value for the top goal, longevity. Details of the inference process are given for the first setting only. They are easily extended to the other settings. The first setting corresponds to an average adult female of Asian race, with little vices or excesses and a reasonable diet. FLARE first infers that:

- Her relative weight is normal (absolute weight ¡ 110 lbs and small frame).
- Her personality type is A, as she is aggressive.
- Her blood pressure is average (normal fat and salt intake).
- Her base longevity is average, namely 67 (range is 48-84).
- Her chances of living longer (i.e., *add* years to base-longevity) are good.

And then, based on this added information, infers that her risk is actually high and though the chances of living longer are good, the actual value added to base-longevity is 0 (i.e., factor = none). Finally, as one would have expected, the woman's longevity is predicted to be average (i.e., 67).

The other settings further illustrate FLARE's ability to perform forward chaining. The second setting corresponds to a very unhealthy older male whose longevity is accordingly predicted to be low and the third setting describes a young healthy female whose life is expectedly predicted to be quite long. Note that though the results may seem impressive, the experiments are only "anecdotal."

| | setting 1 | result 1 | setting 2 | result 2 | setting 3 | result 3 |
|---|---|---|---|---|---|---|
| rel. weight | ? | $normal_1$ | ? | $obese_1$ | ? | $normal_1$ |
| val | ? | $yes_2$ | ? | ? | ? | $yes_2$ |
| heart-dis-risk | ? | ? | ? | $> average_1$ | ? | $average_1$ |
| hddanger | ? | ? | ? | ? | ? | $low_2$ |
| start | yes | yes | yes | yes | yes | yes |
| age | 25-55 | 25-55 | >55 | >55 | <25 | <25 |
| gender | F | F | M | M | F | F |
| base-longevity | ? | $67_1$ | ? | $60_1$ | ? | $72_1$ |
| weight | <110 | <110 | >170 | >170 | 110-170 | 110-170 |
| frame | small | small | small | small | large | large |
| cholesterol | ? | ? | high | high | low | low |
| fat intake | normal | normal | high | high | normal | normal |
| salt intake | normal | normal | high | high | normal | normal |
| blood-pressure | ? | $average_1$ | ? | $> average_1$ | ? | $average_1$ |
| calcium | ? | ? | ? | ? | normal | normal |
| osteoporo-risk | ? | ? | ? | ? | ? | $average_1$ |
| smoker | no | no | yes | yes | no | no |
| outlook | ? | ? | ? | $bleak_2$ | ? | $fair_3$ |
| race | asian | asian | caucasian | caucasian | caucasian | caucasian |
| origin | ? | ? | medit. | medit. | n-amer. | n-amer. |
| risk | ? | $high_2$ | ? | $high_2$ | unknown | unknown |
| personality | aggressive | aggressive | aggressive | aggressive | docile | docile |
| person. type | ? | $type_a_1$ | ? | $type_a_1$ | ? | $type_b_1$ |
| alcohol cons. | moderate | moderate | excessive | excessive | none | none |
| add | ? | $good_1$ | ? | $poor_1$ | ? | $fair_1$ |
| factor | ? | $none_3$ | ? | $minus_12_3$ | ? | $plus_12_4$ |
| longevity | ? | $67_4$ | ? | $48_4$ | ? | $84_5$ |

Table 9: *Health* Knowledge Base

Note that as in classical expert systems, the identification of a closest match during reasoning could be used to extend FLARE so that it may query the user for missing information as well as justify both the queries and the decisions made.

### 3.6 Limitations

The above applications serve to demonstrate that FLARE holds promise. However, FLARE has many important limitations, several of which were mentioned throughout the paper. Some of them are summarized here.

FLARE's use of AVL as a representation language limits its applicability to relatively simple problems. Induction and deduction are carried out within the confines of non-recursive, propositional logic. Such a restriction makes the combination of learning and reasoning more accessible since much research has taken place within this context. However, first-order predicate logic seems a minimum requirement for any system claiming reasoning abilities.

Although FLARE produces good results, the applications it was tested on are relatively simple. For example, many of the databases in the UCI repository have low complexity and relatively unsophisticated learning methods perform well on them. This explains why FLARE's extremely coarse generalization scheme seems sufficient to attain reasonable predictive accuracy. Similarly, the reasoning problems presented are somewhat straightforward. It follows that simple mechanisms such as static priorities and other counting devices used by FLARE are sufficient.

FLARE does not have any meta-level abilities. The system is unable to reason about its own knowledge and is subsequently unable to produce meaningful middle ground solutions. Yet, work on Cyc (Guha & Lenat, 1994) strongly suggests that meta-knowledge is indispensable in carrying out uncertain reasoning.

It is clear that FLARE only "scratches the surface" of the problem of effectively and efficiently combining induction and deduction. Work on ILP (Muggleton, 1992) may shed some light on the issue of bringing systems like FLARE to a first-order logic level.

## 4. Related Work

FLARE follows in the tradition of PDL2 (Giraud-Carrier & Martinez, 1994b) and ILA (Giraud-Carrier & Martinez, 1995), as it attempts to combine inductive learning using prior knowledge together with reasoning. Unlike PDL2 and ILA whose prior knowledge must be pre-encoded and whose reasoning power is limited to classification (i.e. 1-step forward inferences only), FLARE supports the automatic generation of precepts and forward chaining to any arbitrary depth. Whereas PDL2's actual operation tends to decouple learning and reasoning (i.e., the system essentially uses distinct mechanisms to perform either one), ILA implements an inherently more incremental approach by combining them into a 2-phase algorithm that always reasons first and then adapts accordingly. FLARE further extends ILA by providing a natural transformation from constrained first-order clauses to attribute-value vectors and a more accurate characterization of conflicting defaults.

In attempting to construct an unified framework for learning and reasoning, FLARE follows a synergistic approach, similar (at least in concept) to that taken in SOAR (Laird, Newell, & Rosenbloom, 1987) and NARS (Wang, 1993) for example. There are also a variety of inductive learning models and reasoning systems that bear similarity with the corresponding components of FLARE. Some of them are discussed here.

Induction in FLARE is carried out much the same way as in NGE (Salzberg, 1991). However, because generalization is effected only by setting some attribute(s) to don't-care, the produced generalizations or generalized exemplars (Salzberg, 1991), are hyperplanes, rather than hyperrectangles, in the input space. Hence, FLARE implements a nearest-hyperplane learning algorithm. FLARE also uses static and dynamic priorities to break ties between equidistant generalizations. Moreover, where it was shown that overlapping hyperrectangles may hinder performance (Wettschereck & Dietterich, 1994), FLARE allows overlapping hyperplanes for purposes of dealing with conflicting defaults.

In the case that no generalizations are constructed from the training examples, FLARE degenerates into a restricted form of MBR (Stanfill & Waltz, 1986). The distance metric used is similar to IBL's metric (Aha et al., 1991) but it also handles don't-care attributes (which are non-existent in instance-based learners) and treats missing attributes somewhat differently. Where IBL considers missing attributes to be complete mismatches, FLARE chooses a more middle-ground approach that may better capture the inherent notion of missing or "don't-know" attributes.

Learning in FLARE contrasts with algorithms such as CN2 (Clark & Niblett, 1989), where all training examples must be available a priori. Rather, FLARE follows an incremental approach similar to that argued by Elman (1991), except that it is the knowledge itself that is evolved, rather than the system's structure. Moreover, learning in FLARE can be effected continually. Any time an example or a precept is presented and its target output is known, FLARE can adapt.

Prior knowledge may take a variety of forms, some of which are discussed by Mitchell (1980) and Buntine (1990). The form most relevant to FLARE consists of domain-specific inference rules, either pre-encoded or deduced from more general rules. Systems that explicitly combine inductive learning with this kind of prior knowledge include PDLA (Giraud-Carrier & Martinez, 1993), ScNets (Hall & Romaniuk, 1990), ASOCS (Martinez, 1986) and ILP (Muggleton, 1992; Muggleton & De Raedt, 1994). ScNets are hybrid symbolic, connectionist models that aim at providing an alternative to knowledge acquisition from experts. Known rules may be pre-encoded and new rules can be learned inductively from examples. The representation lends itself to rule generation but the constructed networks are complex and generalization does not appear trivial. ASOCS and PDLA are dynamic, self-organizing networks that learn, incrementally, from both examples and rules. In ASOCS, order matters and conflicts are simply solved by giving priority to the most recent rules. PDLA is less order-dependent and provides evidence-driven mechanisms for the handling of conflicts. As in ScNets, prior knowledge in ASOCS and PDLA takes the form of explicitly encoded, domain-specific rules. FLARE's approach is more flexible. Because the system can reason, domain-specific rules (or precepts) can be deduced automatically from more general rules. ILP models offer the same flexibility. At the intersection of logic programming and inductive learning, ILP takes advantage of the full expressiveness of first-order predicate logic to learn first-order theories from background theories and examples. FLARE's representation language, though capable of handling both nominal and linear (including continuous and numerical) data, is only as expressive as non-recursive, propositional clauses. However, in this simpler setting, FLARE supports evidential reasoning and the prioritization of rules.

FLARE's use of rules and similarity in reasoning is similar to CONSYDERR's (Sun, 1992). However, CONSYDERR is strictly concerned with a connectionist approach to concept representation and commonsense reasoning. The resulting model is elegant. It consists of a two-level architecture that naturally captures the dichotomy between concepts and the features used to describe them. However, it does not address the problem of learning (how such a skill could be incorporated is also unclear) and is currently limited to reasoning from concepts. FLARE's representation is not as elegant but the model can effectively reason from concepts or from features. CONSYDERR deals only with Boolean features and a concept's representation is limited to a single conjunction of features. FLARE's concepts generally consist of several conjunctions of features, each representing partial and complementary definitions of the concept. Also, since the domain of features is not restricted, FLARE uses a more general distance metric than CONSYDERR's similarity measure based on feature overlap. However, FLARE currently has no mechanisms for individual weighting of features, which may cause performance degradation and increased memory requirements in the presence of a large number of irrelevant features.

FLARE's ability to evolve its knowledge base over time is similar to that found in theory-refinement systems such as RTLS (Ginsberg, 1990), EITHER (Ourston & Mooney, 1990, 1994) and KBANN (Towell, Shavlik, & Noordewier, 1990; Towell & Shavlik, 1994). RTLS implements a 3-phase algorithm for refinement. It first reduces the current theory to a form suitable for inductive learning, then performs learning and, finally, retranslates the result into a new theory. This process is potentially costly. In FLARE, the language of the theory is the same as the language of induction, that is, the theory is always in reduced form. Though the language is not as rich, it allows revision to take place efficiently for each new example, incrementally. EITHER is similar to FLARE as it assumes an approximate theory and allows correction of both overly-general and overly-specific rules. The mechanisms for revision are different. EITHER may add/remove antecedents and rules, while FLARE may remove antecedents and add rules and exceptions. EITHER currently only handles Boolean attributes, while FLARE has no such restriction. However, EITHER uses both explanation-based learning and inductive learning in revision, while FLARE is strictly inductive. KBANN, like EITHER, only deals with propositional, non-recursive Horn clauses. Prior knowledge is expressed identically to FLARE's pre-encoded precepts (i.e., domain-specific inference rules in the form of Prolog-like clauses). KBANN translates the given knowledge base into an equivalent artificial neural network (ANN) and may then perturb it and learn using the backpropagation algorithm. In FLARE, there is no ANN; the knowledge base is simply stored as individual rules. Overall, FLARE provides a slightly more general and synergistic approach. New evidence is constantly used to revise the current state of knowledge. There are currently no mechanisms in FLARE to deal explicitly with fuzzy rules. However, several mechanisms exist to handle inconsistencies and conflicts. FLARE always makes a decision based on available evidence. A confidence level can also be produced to characterize the "goodness" of the decision.

FLARE's limited handling of non-monotonicity differs from the approach taken in logic. Non-monotonic logics typically extend first-order predicate logic through added "machinery," such as circumscription (McCarthy, 1980), semi-normal defaults (Reiter & Griscuolo, 1981) or hierarchical theories (Konolige, 1988), while essentially preserving consistency. FLARE's approach consists of tolerating inconsistencies in the knowledge base but providing reasoning mechanisms that ensure that no inconsistent conclusions are ever reached. It essentially consists of using normal defaults for inheritance and an external criterion for cancellation (Vilain, Koton, & Chase, 1990). The current criterion relies mostly on a simple counting argument (for dynamic priorities and covers). Though this approach has proven sufficient for the simple propositional examples described here, it is likely to break down on more sophisticated examples and domains.

## 5. Conclusion

This paper highlights some of the interdependencies between learning and reasoning and details a system, called FLARE, that combines inductive learning using prior knowledge together with reasoning within the confines of non-recursive, propositional logic. Several important positive conclusions may be drawn from the results of this research. In particular,

- Performance in induction is improved in terms of both memory requirement and generalization when prior knowledge is used.
- Induction from examples can be used to effectively resolve conflicting defaults extensionally.
- Combining rule-based and similarity-based reasoning provides a useful means of performing approximate reasoning and tends to reduce brittleness.
- Induction offers a valuable complement to classical knowledge acquisition techniques from experts.

Experiments with FLARE on a variety of applications demonstrate promise. However, much work still remains to be done to achieve a more complete and meaningful integration of learning and reasoning. Areas of future work include the following:

- Designing mechanisms to use reasoning to guide learning.
- Attempting to overcome (or appropriately use) the order-dependency.
- Providing support for internal disjunction.
- Improving the use of inductively learned rules in reasoning (the support is available but the induction may not produce useful rules).
- Possibly incorporating backward chaining.
- Translating the system's knowledge base back from AVL to FOL.
- Further experimenting with larger applications.
- Extending the language to first-order.

## Acknowledgements

This work was supported in part by grants from Novell Inc. and WordPerfect Corp. Many thanks also to our reviewers for helpful and constructive comments.